\documentclass[10pt,journal,compsoc]{IEEEtran}
\usepackage{times}
\usepackage[nocompress]{cite}
\usepackage{amsmath,amsthm,amsfonts,amssymb,times}
\usepackage{mathtools}
\usepackage{acronym}
\usepackage[linesnumbered,ruled,vlined]{algorithm2e}
\usepackage{algpseudocode}
\usepackage{algcompatible}

\usepackage{array}
\usepackage{tabularx}
\usepackage{multirow}
\usepackage{setspace}

\usepackage{mdwmath}
\usepackage{mdwtab}
\usepackage{eqparbox}

\usepackage{graphicx}
\usepackage{subfigure}
\usepackage{psfrag}

\usepackage{url}

\usepackage{flushend}
\usepackage{here}
\usepackage{bm}
\usepackage{textcomp}
\usepackage{xcolor}
\usepackage{physics}
\usepackage{hhline}

\selectfont

\newcommand{\fracslant}[2]{
	\raisebox{0.4ex}{\small $#1$}
	\raisebox{0ex}{\large $/$}
	\raisebox{-0.2ex}{\small $#2$}
}
\newcommand{\mtscript}[1]{\mbox{\scriptsize\em{#1}}}

\newtheorem{proposition}{Proposition}
\newtheorem{define}{Definition}

\begin{document}

\title{Acceleration Method for Learning\\
	Fine-Layered Optical Neural Networks}
\author{Kazuo~Aoyama and Hiroshi~Sawada 
\IEEEcompsocitemizethanks{\IEEEcompsocthanksitem K. Aoyama and H. Sawada 
are with 
NTT Communication Science Laboratories, 
2-4, Hikaridai, Seika-cho, Soraku-gun, Kyoto 619-0237 Japan.\protect\\
E-mail: \{kazuo.aoyama.rd, hiroshi.sawada.wn\}@hco.ntt.co.jp 
}
}

\IEEEtitleabstractindextext{%
\begin{abstract}
An optical neural network (ONN) is a promising system due to 
its high-speed and low-power operation.
Its linear unit performs a multiplication of an input vector and 
a weight matrix in optical analog circuits. 
Among them, a circuit with a multiple-layered structure of 
programmable Mach-Zehnder interferometers (MZIs) can realize 
a specific class of unitary matrices with a limited number of MZIs 
as its weight matrix.
The circuit is effective for balancing
the number of programmable MZIs and ONN performance.
However, it takes a lot of time to learn MZI parameters 
of the circuit with a conventional automatic differentiation (AD),
which machine learning platforms are equipped with.
To solve the time-consuming problem, 
we propose an acceleration method for learning MZI parameters.
We create customized complex-valued derivatives 
for an MZI, exploiting Wirtinger derivatives and a chain rule.
They are incorporated into
our newly developed function module implemented in C++ 
to collectively calculate their values in a multi-layered structure.
Our method is simple, fast, and versatile  
as well as compatible with the conventional AD.
We demonstrate that 
our method works 20 times faster than the conventional AD 
when a pixel-by-pixel MNIST task is performed in 
a complex-valued recurrent neural network 
with an MZI-based hidden unit.
\end{abstract}

}
\maketitle

\IEEEpeerreviewmaketitle

\ifCLASSOPTIONcompsoc
\IEEEraisesectionheading{\section{Introduction}\label{sec:intro}}
\else
\section{Introduction}\label{sec:intro}
\fi
Optical neural networks (ONNs) have attracted much attention
because of their high speed and extremely low power consumption
\cite{harris,lin-ozcan,delima,bogaerts,wetzstein,xu},
compared with conventional digital computer systems.
ONNs process information encoded by amplitude and phase of light waves
in passive analog circuits exploiting optical phenomena such as 
transmission, resonance, interference, and diffraction.
Such optical analog circuits operate in principle without energy.
Linear units in ONNs perform analog multiplication of 
a weight matrix and 
an input vector that is a signal of encoded information.
In the multiplication exploiting the interference 
\cite{shen,lin-ozcan,hamerly,hughes}, 
signals and weights are regarded as complex numbers.
Since the energy is preserved in the vector-matrix multiplication,
the weight matrix is a unitary matrix.

Among such linear units,
there are ones with a unitary matrix implemented in
programmable Mach-Zehnder interferometers (MZIs)
\cite{shen,fang,pai}.
The linear unit using MZIs has two main advantages for
designing ONNs:
One is that 
the ONNs with the linear units can learn 
not a unitary matrix itself but parameters of the MZIs directly 
\cite{jing}.
The other is that 
the ONNs can balance their performance and the limited number
of their MZIs by using the characteristic that 
not only full-capacity unitary matrix but also its specific class 
is realized by MZI-based matrices \cite{fang}.
For these advantages,
we focus on the linear units using MZIs
and discuss methods for learning their parameters.

The learning method \cite{jing} 
can generate specific classes of unitary matrices 
using fewer parameters 
in an MZI-based matrix 
than those required for realizing any unitary matrix. 
However, this method requires a lot of computational cost of elapsed time 
to learn the parameters.
Two main factors make this method costly.
One is to represent a weight matrix by a product of several MZI-representation matrices.
This is equivalent to adopting a sequence of several linear units,
resulting in a deeper neural network.
The other is to use the conventional automatic differentiation (AD) \cite{baydin} 
without any change,
which machine learning frameworks such as 
TensorFlow \cite{tensorflow} and PyTorch \cite{pytorch} 
are equipped with. 
The machine learning framework requires more computational time for 
the deeper neural networks.
Besides, we need flexible representations for an MZI 
to systematically deal with 
various MZI-representation matrices \cite{jing,fang,pai,bogaerts}.

To solve the problems,
we propose an acceleration learning method to remove the foregoing factors and 
adapt to the various MZI-representation matrices, 
retaining the compatibility with the conventional AD.
Our proposed learning method is simple, fast, versatile, and easy-to-use.
%
Our method is based on three newly developed techniques.
The first is to prepare two constituent unitary matrices corresponding to
two basic components of an MZI, a phase shifter (PS) and 
a directional coupler (DC) (or beam splitter) \cite{shen,fang,pai},
and represent an MZI matrix by the combination of representation matrices of 
the PS and the DC.
This leads to the versatility of our method and makes a matrix formulation simple.
The second is to create customized complex-valued derivatives for matrices of 
pairs of the PS and the DC, i.e., PSDC and DCPS,
by exploiting Wirtinger derivatives and a chain rule \cite{hjorungnes}.
The last is to develop a function module implemented in C++ 
to collectively calculate values required in learning of 
a linear unit with the several combination matrices.
By incorporating the customized derivatives to the function module, 
our method achieves fast learning.

Our contributions are threefold:
\begin{enumerate}
\item We present a fine-layered linear unit where 
a weight matrix is represented by a product of structured unitary matrices 
implemented in phase shifters (PS) and directional couplers (DC).
Owing to this representation, 
we can simply formulate a weight matrix and 
easily modify it according to various MZI implementations.
\item We propose an acceleration learning method for a fine-layered linear unit. 
We create customized complex-valued derivatives for products of 
the structured matrices, 
exploiting a chain rule and Wirtinger derivatives.
The customized derivatives are incorporated to our newly developed function module 
implemented in C++ 
to collectively calculate their values. 
Since our proposed method is compatible with automatic differentiation (AD),
we can easily use it in the conventional machine learning platforms.
\item We demonstrate that 
our proposed method works
20 times faster than the conventional AD
when a pixel-by-pixel MNIST task \cite{jing} is performed in a complex-valued 
recurrent neural network where a hidden unit is a fine-layered linear unit 
based on the PSDC.
\end{enumerate}

The remainder of this paper consists of the following six sections.
Section \ref{sec:relate} briefly reviews related work.
Section \ref{sec:mzi} describes unitary matrices represented by MZIs.
Section \ref{sec:math} provides background knowledge on learning 
a complex-valued linear unit. 
Section \ref{sec:prop} explains our learning method in detail.
Section \ref{sec:exp} shows our experimental settings and demonstrates the results.
The final section provides our conclusion and future work.

\section{Related Work}\label{sec:relate}
This section reviews two topics 
regarding how to determine parameters of programmable MZIs in a unitary matrix.

\subsection{Constructing Linear Units By Using MZIs }\label{subsec:mit}
A method in \cite{shen} constructs a linear unit with a weight matrix
implemented in programmable MZIs.
What is learned with the method is 
neither a unitary matrix nor parameters of programmable MZIs 
but a weight matrix itself.
It first obtains optimized weight matrix $W$ by learning a weight matrix 
with a conventional algorithm (e.g., \cite{hinton}).
$W$ is decomposed with singular value decomposition (SVD) 
as $W\!=\!U \Sigma V^{\dagger}$,
where $U$ denotes a unitary matrix,
$V^{\dagger}$ denotes the conjugate transpose of unitary matrix $V$,
and $\Sigma$ denotes a rectangular diagonal matrix.
The unitary matrices $U$ and $V^{\dagger}$ are implemented in
programmable MZIs 
by a triangular-structure implementation method
\cite{reck,miller}.
The diagonal matrix $\Sigma$ can be implemented 
in optical attenuators and phase shifters.
Instead of the triangular-structure implementation method,
we can employ a rectangular-structure method \cite{clements,pai}. 
Although this scheme can implement any unitary matrix in MZIs, 
it can not generate a specific class of unitary matrices 
by fewer MZIs. 
This is a problem when designing a higher-performance ONN 
using limited physical resources.

\renewcommand{\arraystretch}{1.1}
\setlength\doublerulesep{0.5pt}
\begin{table}[t]
\caption{Learning Methods for Unitary Matrices}
\begin{center}
\vspace*{-2mm}
\begin{tabular}{|c||c|c|c|c|}\hline
Constraint & Optimization & \multicolumn{3}{c|}{Structure}\\ \hline
\multirow{3}{*}{\shortstack{Matrix\\representation\\capacity}} &
 \multicolumn{2}{c|}{\multirow{3}{*}{\shortstack{Full unitary}}} & 
 \multicolumn{2}{c|}{Specified class}\\\cline{4-5}
 & \multicolumn{2}{c|}{~}& \multirow{2}{*}{Fixed} 
 & \multirow{2}{*}{\shortstack{Variable\\(to full)}}\\
 & \multicolumn{2}{c|}{~} & & \\ \hhline{=====}
\multirow{2}{*}{Method} & \cite{bansal}, \cite{wisdom} & 
\cite{maduranga} & \cite{mathieu} & \cite{mhammedi},\cite{jing} \\
 & \cite{vorontsov},\cite{wolter} & & \cite{arjovsky} & Ours\\\hline
\end{tabular}
\label{table:method}
\vspace*{-3mm}
\end{center}
\end{table}
\renewcommand{\arraystretch}{1.0}

\subsection{Learning Methods For Unitary Matrices}\label{subsec:learn_unitary}
Neural networks with unitary matrices as 
their weight matrices
have been studied 
to alleviate a problem of vanishing or exploding gradients in 
weight optimization \cite{arjovsky,wisdom,mhammedi}.
Table~\ref{table:method} summarizes
learning methods for unitary matrices.
There are two types of constraints for generating a unitary matrix,
optimization and structural constraints.
A generated unitary matrix has a distinct representation capacity
from a fixed specified class to a full-capacity unitary representation.

In the methods based on optimization constraints,
a convenient one is to add the constraint to a loss function as a regularizer 
\cite{bansal}.
A more strict method is to optimize a weight matrix along Stiefel manifold
whose tangent spaces are endowed with a Riemannian metric, 
using geodesic gradient descent \cite{wisdom,vorontsov,wolter}.
These methods can generate a full-capacity unitary matrix.
By contrast, 
the methods based on structural constraints generate a unitary matrix 
expressed by a product of structured unitary matrices
such as Givens rotation \cite{mathieu,jing}, 
Householder reflection \cite{arjovsky,mhammedi},
and skew-Hermitian matrices \cite{maduranga}.
Depending on the constituent structured unitary matrices and 
their parameterization,
the generated matrix has unique representation capacity.
Unitary matrices by the methods \cite{mhammedi},\cite{jing} vary their capacity
from a specified-class to full unitary
while those \cite{mathieu},\cite{arjovsky} have fixed and restricted capacity.

From the viewpoint of optical-circuit implementations,
the optimization-constraint approach has 
a serious problem that 
it is difficult to obtain an exact unitary matrix.
The structure-constraint approach has an advantage of 
being able to construct an exact unitary matrix. 
In particular, the method \cite{jing}, 
which is suitable to the MZI implementation of a unitary matrix, 
can generate specific classes of unitary matrices 
using fewer parameters in structured matrices.
However, this method needs a lot of computational time 
for learning the MZI parameters.

Thus the previous methods have some problems to realize linear units in ONNs 
with unitary matrices implemented in 
the MZIs.
Our proposed learning method in Table~\ref{table:method} 
solves the problem that the method in \cite{jing} has 
and furthermore improves versatility of representation matrices of the MZIs.

\section{Unitary Matrices Represented By MZIs}\label{sec:mzi}
We show that an MZI is represented by various unitary matrices 
depending on its structure
and describe that any $n\!\times\! n$ unitary matrix is realized 
by a product of MZI-representation matrices and a diagonal
unitary matrix.

\subsection{Representation Matrix of MZI}\label{subsec:mzi}
We first define a unitary matrix.
Let $a_{(n)}$ denote an element of the $n$-dimensional unitary group $U(n)$
and $A_{(n)}$ denote the representation matrix of $a_{(n)}$.
Then $n\!\times\! n$ unitary matrix $A_{(n)}$ satisfies 
the unitary constraints of $A_{(n)}A_{(n)}^{\dagger}\!=\!I$ and 
$\left\vert \mathrm{det} A_{(n)}\right\vert\!=\!1$, where
$I$ denote the $n\!\times\! n$ identity matrix.
When $n\!=\! 2$,
any $2\!\times\! 2$ unitary matrix $A_{(2)}$ has 
four independent real-number parameters 
corresponding to the degree of freedom of the unitary group $U(2)$.
\begin{figure}[t]
\begin{center}
	\subfigure[{\normalsize MZI}]{
		\psfrag{X}[c][c][1.0]{$x_1$}
		\psfrag{Y}[c][c][1.0]{$x_2$}
		\psfrag{Z}[c][c][1.0]{$y_1$}
		\psfrag{U}[c][c][1.0]{$y_2$}
		\psfrag{V}[c][c][1.0]{$(\phi,\theta)$}
		\includegraphics[width=30mm]{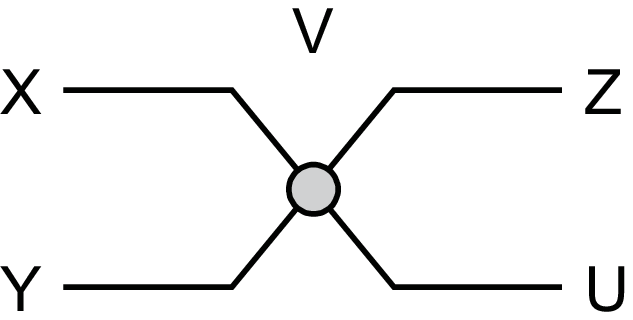}
	} \hspace*{1mm}
	\subfigure[{\normalsize (PSDC)$^2$}]{
		\psfrag{A}[c][c][1.0]{$x_1$}
		\psfrag{B}[c][c][1.0]{$x_2$}
		\psfrag{C}[c][c][1.0]{$y_1$}
		\psfrag{D}[c][c][1.0]{$y_2$}
		\psfrag{E}[c][c][.9]{PS$_1$}
		\psfrag{F}[c][c][.9]{DC$_1$}
		\psfrag{G}[c][c][.9]{PS$_2$}
		\psfrag{H}[c][c][.9]{DC$_2$}
		\psfrag{P}[c][c][.9]{$\phi$}
		\psfrag{T}[c][c][.9]{$\theta$}
		\includegraphics[width=42mm]{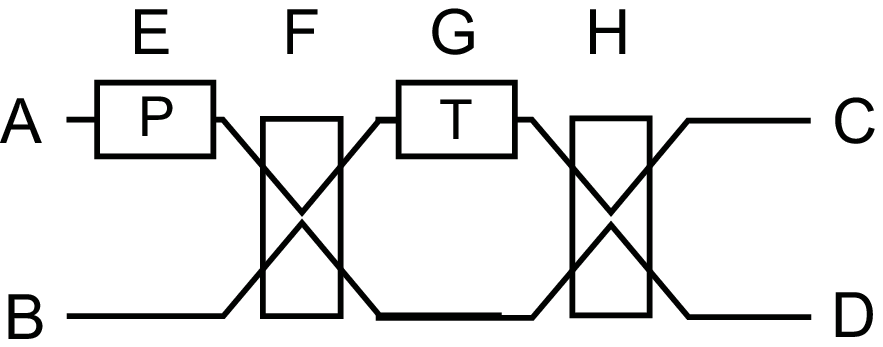}
	}
\end{center}
\vspace*{-2mm}
\caption{Symbol of (a) MZI and structure of (b) (PSDC)$^2$.
$x_1,~x_2\!\in\!{\mathbb C}$ and $y_1,~y_2\!\in\!{\mathbb C}$ denote
input and output complex values and
$\phi$ and $\theta$ are real-valued parameters.}
\label{fig:mzi}
\end{figure}

An MZI is a two-port optical circuit in Fig.~\ref{fig:mzi}(a)
and linearly transforms 
complex-valued input vector $(x_1,x_2)^T$ to output vector $(y_1,y_2)^T$ 
by a variable transformation matrix with two parameters of $\phi$ and $\theta$,
where $(x_1,x_2)^T$ denotes the transpose of $(x_1,x_2)$, i.e.,
the input vector is represented as the column vector.
The MZI consists of two programmable phase shifters (PS) and 
two directional couplers (DC) with a fixed 0.5:0.5 power split ratio.
We adopt a symbol for an MZI shown in Fig.~\ref{fig:mzi}(a) 
and illustrate a typical structure for the MZI, 
which is the serial connection of the PS and the DC, 
i.e., (PSDC)(PSDC) or (PSDC)$^2$ in Fig.~\ref{fig:mzi}(b). 
Representation matrices of the programmable PS and the fixed DC, 
$M_{[{\mtscript PS}(\phi)]}$ and $M_{[{\mtscript DC}]}$, 
are expressed as 
\begin{equation}
M_{[{\mtscript PS}(\phi)]} = 
\begin{pmatrix}e^{i\phi} &0\\0 &1\end{pmatrix}\:,
\hspace*{3mm}
M_{[{\mtscript DC}]} = \frac{1}{\sqrt{2}}\begin{pmatrix}1&i\\i&1\end{pmatrix}\: ,
\label{eq:mat_ps_dc}
\end{equation}
where $0\!\leq\!\phi\!\leq\! 2\pi$ and $i^2\!=\!-1$.
Since $M_{[{\mtscript PS}(\phi)]}$ and $M_{[{\mtscript DC}]}$ satisfy 
the unitary constraints, a product of the PS- and the DC-representation 
matrices becomes a unitary matrix.

An actual transformation matrix depends on the connection of PS's and DC's.
For instance, Fang's matrix ($R_F$) \cite{fang} corresponding to the structure 
in Fig.~\ref{fig:mzi}(b) is expressed by
\begin{align}
R_F &= M_{[{\mtscript DC}]}\, M_{[{\mtscript PS}(\theta)]}\,
M_{[{\mtscript DC}]}\, M_{[{\mtscript PS}(\phi)]}\notag\\
&= ie^{i\frac{\theta}{2}}\begin{pmatrix}
e^{i\phi}\sin\frac{\theta}{2} & \cos\frac{\theta}{2}\\
e^{i\phi}\cos\frac{\theta}{2} & -\sin\frac{\theta}{2}\end{pmatrix}
=\frac{1}{2}\begin{pmatrix}
e^{i\phi}\beta & i\alpha\\ ie^{i\phi}\alpha & -\beta \end{pmatrix}\notag\:,\\
& \hspace{31mm} \alpha = e^{i\theta}+1, \hspace{3mm} \beta = e^{i\theta}-1 \: .
  \label{eq:fang_mat}
\end{align}
Pai's matrix ($R_P$) \cite{pai} representing (DCPS)(DCPS) is 
the transpose matrix of $R_F$ as 
\begin{equation}
R_P = M_{[{\mtscript PS}(\theta)]}\, M_{[{\mtscript DC}]}\, 
M_{[{\mtscript PS}(\phi)]}\, M_{[{\mtscript DC}]} = R_F^T \:.
\label{eq:pai_mat}
\end{equation}
Besides, a matrix ($R_M$) for (DCPS)(PSDC) is expressed by
\begin{align}
R_M &= M_{[{\mtscript DC}]}\, M_{[{\mtscript PS}(\theta)]}\,
M_{[{\mtscript PS}(\phi)]}\, M_{[{\mtscript DC}]}\notag\\
&= \frac{1}{2}\begin{pmatrix}
e^{i\phi}\!-\!e^{i\theta} & i(e^{i\phi}\!+\!e^{i\theta}) \\
i(e^{i\phi}\!+\!e^{i\theta}) & -(e^{i\phi}\!-\!e^{i\theta})\end{pmatrix}
\label{eq:m_mat}
\end{align}
An MZI with two parameters is represented with three distinct matrices of 
$R_F$, $R_P$, and $R_M$ 
if each of the two phases $\phi$ and $\theta$ is regarded as relative phase
and its initial phase difference is ignored.
Thus there are various MZI-representation matrices. 

When selecting one of the three MZI-representation matrices,
e.g., $R_F$,
we can realize any $2\!\times\! 2$ unitary matrix 
by Clements' method \cite{clements} as 
\begin{equation}
A_{(2)} = D\cdot R_F\, ,\hspace{3mm}
D = \begin{pmatrix}e^{i\delta_0}&0\\0&e^{i\delta_1}\end{pmatrix}\: ,
\label{eq:A2_DR}
\end{equation}
where $0\!\leq\!\delta_0,~\delta_1\!\leq\! 2\pi$.
$A_{(2)}$ with four parameters is expressed by a product of 
a diagonal unitary matrix with two parameters and an MZI-representation matrix
with two parameters.

\subsection{Product of MZI-Representation Matrices}\label{subsec:prod}
Using the $2\!\times\! 2$ MZI-representaion matrix,
we consider an $n\!\times\! n$ unitary matrix corresponding to 
an $n$-port optical circuit based on MZIs.
Let us define $n\!\times\! n$ unitary matrix $T_{(p,q:n)}$ represented 
by a single MZI as
\begin{equation}
{\normalsize T_{(p,q:n)}} \coloneqq \begin{pmatrix}
1 &0 &\cdots &\cdots &0 &0\\
0 &1 & & & &0\\
\vdots & & {\normalsize w_{pp}} & {\normalsize w_{pq}} & &\vdots\\
\vdots & & {\normalsize w_{qp}} & {\normalsize w_{qq}} & &\vdots\\
\vdots & & & &1 &0\\
0 &0 &\cdots &\cdots &0 &1 \end{pmatrix}\: ,
\label{eq:t_mat}
\end{equation}
where $p\!<\!q\!\leq\!n\!\in\!{\mathbb Z}$ and 
$w_{pq}\!\in\!{\mathbb C}$ denotes the $p$th-row and $q$th-column element.
The others, $w_{pp}$, $w_{qp}$, and $w_{qq}$, denote the elements 
according to the same rule.
The elements of $w_{pp}$, $w_{pq}$, $w_{qp}$, and $w_{qq}$ correspond to 
$w_{11}$, $w_{12}$, $w_{21}$, and $w_{22}$, of 
a $2\!\times\!2$ MZI-representation matrix, respectively.
Then $n\!\times\! n$ matrix $T_{(p,q:n)}$ has 
two independent real-valued parameters such as $\phi$ and $\theta$.
For instance, when $R_F$ in Eq.~(\ref{eq:fang_mat}) is used,
$w_{pp}\!=\! e^{i\phi}\beta/2$, $w_{pq}\!=\! i\alpha/2$,
$w_{qp}\!=\! ie^{i\phi}\alpha/2$, $w_{qq}\!=\! -\beta/2$.

\begin{figure}[t]
\begin{center}
	\psfrag{A}[r][r][1.0]{$x_1$} \psfrag{B}[r][r][1.0]{$x_2$}
	\psfrag{C}[r][r][1.0]{$x_3$} \psfrag{D}[r][r][1.0]{$x_4$}
	\psfrag{E}[l][l][1.0]{$y_1$} \psfrag{F}[l][l][1.0]{$y_2$}
	\psfrag{G}[l][l][1.0]{$y_3$} \psfrag{H}[l][l][1.0]{$y_4$}
	\psfrag{R}[l][l][0.95]{} \psfrag{S}[l][l][0.95]{}
	\psfrag{T}[l][l][0.95]{} \psfrag{U}[l][l][0.95]{}
	\psfrag{K}[c][c][1.1]{${\bm x}$}
	\psfrag{L}[c][c][1.0]{$S_{A1}$} \psfrag{M}[c][c][1.0]{$S_{B1}$}
	\psfrag{N}[c][c][1.0]{$S_{A2}$} \psfrag{O}[c][c][1.0]{$S_{B2}$}
	\psfrag{P}[c][c][1.0]{$D$} 
	\psfrag{Q}[c][c][1.1]{${\bm y}$}
	\psfrag{d}[c][c][.7]{} \psfrag{b}[l][l][.7]{}
	\psfrag{c}[c][c][.7]{} \psfrag{e}[l][l][.7]{}
	\psfrag{p}[c][c][.7]{} \psfrag{q}[l][l][.7]{}
	\psfrag{k}[c][c][.7]{} \psfrag{h}[l][l][.7]{}
	\psfrag{u}[c][c][.7]{} \psfrag{v}[l][l][.7]{}
	\psfrag{s}[c][c][.7]{} \psfrag{x}[l][l][.7]{}
	\includegraphics[width=70mm]{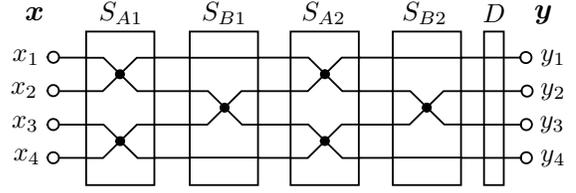}
\end{center}
\vspace*{-2mm}
\caption{
Rectangular structure consisting of
a product of block-diagonal matrices ($S_{A1},S_{B1},S_{A2},S_{B2}$) and 
one diagonal unitary matrix ($D$)
that realizes any $4\!\times\! 4$ unitary matrix.
}
\label{fig:clements}
\vspace*{-3mm}
\end{figure}

Any $n\!\times\!n$ unitary matrix is decomposed to
$n(n-1)/2$ $T_{(p,q:n)}$'s 
and a single $n\!\times\!n$ diagonal unitary matrix $D$ with $n$ parameters 
by Clements' method \cite{clements}. 
The method sequentially determines two parameters of each $T_{(p,q:n)}$ and 
$n$ parameters of the $D$ 
by the procedure similar to Gaussian elimination.
By changing the order of commutative matrices in the obtained $T_{(p,q:n)}$'s,
the method generates a product of unitary matrices $S$ 
with a regular rectangular structure.
An example of the commutative matrices $T_{(p,q:n)}$'s is shown as follows.
$4\!\times\!4$ unitary matrix 
$S_{((1,2),(3,4):4)}\!=\!T_{(1,2:4)}T_{(3,4:4)}\!=\!T_{(3,4:4)}T_{(1,2:4)}$.

Figure~\ref{fig:clements} shows a diagram of a linear unit with 
the rectangular structure generated by Clements' method, 
which realizes any $4\!\times\!4$ unitary matrix.
The foregoing product $S_{((1,2),(3,4):4)}$ is $S_{A1}$ 
and $S_{(2,3):4}\!=\! T_{(2,3:4)}$ is $S_{B1}$. 
Given $R_F$ in Eq.~(\ref{eq:fang_mat}) as the MZI-representation matrix,
$S_{A1}\!=\!S_{((1,2),(3,4):4)}$ and  
$S_{B1}\!=\!S_{((2,3):4)}$ are expressed by
\begin{align}
S_{A1} &=\frac{1}{2}
\begin{pmatrix}
e^{i\phi_{1}}\beta_{1} & i\alpha_{1} & 0 & 0\\
ie^{i\phi_{1}}\alpha_{1} & -\beta_{1} & 0 & 0\\
0 & 0 & e^{i\phi_{2}}\beta_{2} & i\alpha_{2} \\
0 & 0 & ie^{i\phi_{2}}\alpha_{2} & -\beta_{2} \end{pmatrix}\:, 
\label{eq:sa1}\\
S_{B1} &=\frac{1}{2}
\begin{pmatrix}
2 & 0 & 0 & 0 \\
0 & e^{i\phi_3}\beta_3 & i\alpha_3 & 0\\
0 & ie^{i\phi_3}\alpha_3 & -\beta_3 & 0 \\
0 & 0 & 0 & 2 \end{pmatrix}\: . \label{eq:sb1}
\end{align}
In the rectangular structure for realizing an $n\!\times\!n$ unitary matrix,
$S_{A(L)}$ and $S_{B(L)}$, which we term A-type and B-type fine layers, 
alternately align.
$S_{A(L)}$ and $S_{B(L)}$
have $\lfloor n/2\rfloor$ and $\lfloor (n-1)/2\rfloor$ MZIs.
Regarding $L$, 
$L\!=\!\lceil n/2 \rceil$ for $S_{A(L)}$ and 
$L\!=\!\lfloor n/2 \rfloor$ for $S_{B(L)}$, $n\!\geq\! 3$, i.e., 
the total number of fine layers is $n$ 
for $n\!\geq\!3$ and $1$ for $n\!=\!2$.

In addition to the structural regularity,
the rectangular structure has an advantage that
we can vary matrix representation capacity with the number of fine layers 
from a specific class to a full-capacity unitary matrix.
This characteristic allows us to control linear-unit performance 
by the number of optimized parameters, 
which corresponds to the number of MZIs in physical resources.
For this reason, 
we select a unitary matrix with the rectangular structure 
for our proposed learning method.

\vspace*{-1ex}
\section{Learning Complex-Valued Linear Units}\label{sec:math}

\begin{figure}[t]
\begin{center}
	\subfigure[{\normalsize Forward}]{
		\psfrag{A}[c][c][0.9]{$x_1$}
		\psfrag{B}[c][c][0.9]{$x_2$}
		\psfrag{C}[c][c][0.9]{$y_1$}
		\psfrag{D}[c][c][0.9]{$y_2$}
		\psfrag{P}[c][c][0.9]{$w_{11}$}
		\psfrag{Q}[l][l][0.9]{$w_{12}$}
		\psfrag{R}[l][l][0.9]{$w_{21}$}
		\psfrag{S}[c][c][0.9]{$w_{22}$}
		\includegraphics[width=31mm]{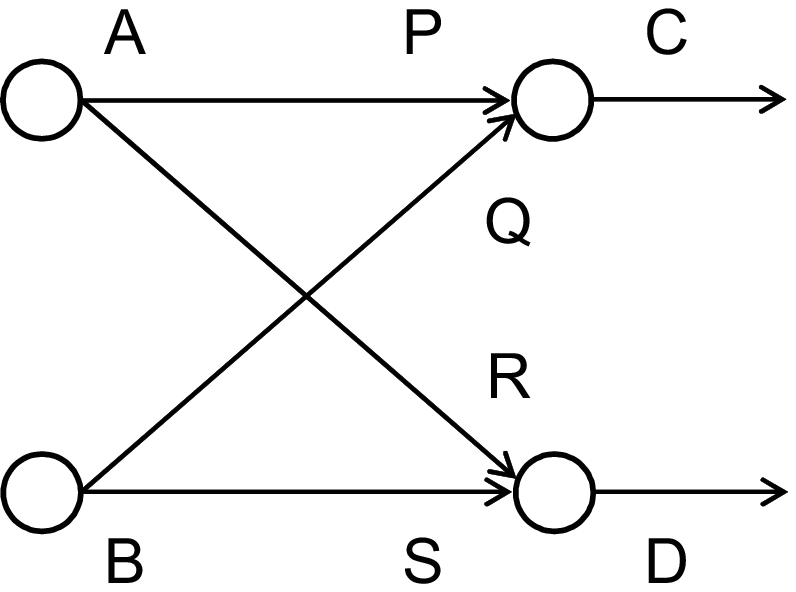}
	} \hspace*{6mm}
	\subfigure[{\normalsize Backward}]{
		\psfrag{A}[r][r][1.0]{$\frac{\partial{\cal L}}{\partial x_1}$}
		\psfrag{B}[c][c][1.0]{$\frac{\partial{\cal L}}{\partial x_2}$}
		\psfrag{C}[l][l][1.0]{$\frac{\partial{\cal L}}{\partial y_1}$}
		\psfrag{D}[c][c][1.0]{$\frac{\partial{\cal L}}{\partial y_2}$}
		\psfrag{P}[l][l][0.9]{$w_{11}$}
		\psfrag{Q}[c][l][0.9]{$w_{21}$}
		\psfrag{R}[c][l][0.9]{$w_{12}$}
		\psfrag{S}[l][l][0.9]{$w_{22}$}
		\includegraphics[width=31mm]{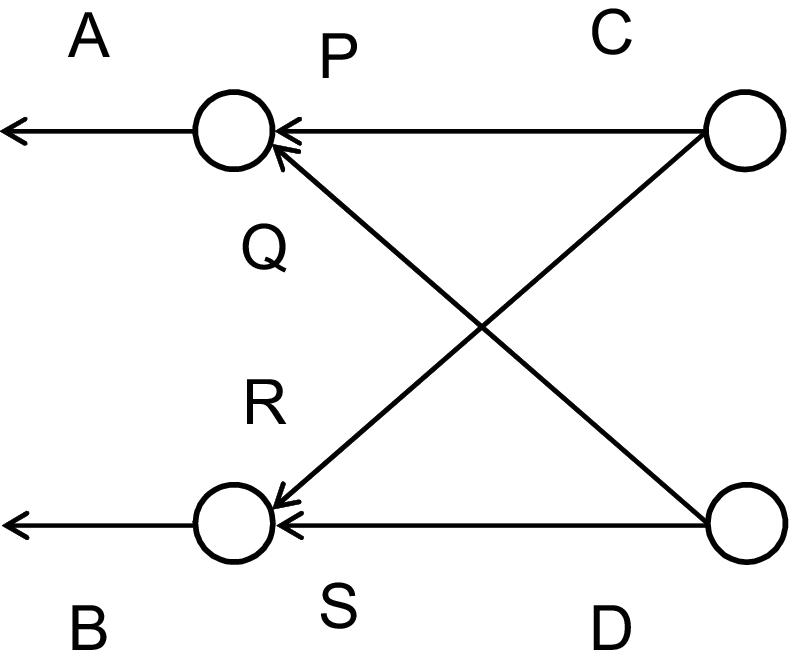}
	}
\end{center}
\vspace*{-2mm}
\caption{Diagram representing the relationship between inputs and outputs 
for (a) forward and (b) backward process in a generic linear unit.}
\label{fig:fwbkmodel}
\end{figure}

We begin by a generic real-valued linear unit 
to review learning of linear units based on 
the backpropagation algorithm with gradient descent and 
automatic differentiation (AD). 
Then we extend the real-valued AD to complex-valued one using Wirtinger derivatives.

\vspace*{-1ex}
\subsection{Learning Linear Units and Auto-Differentiation}\label{subsec:ad}
Automatic differentiation (AD) 
is equipped with most of machine learning platforms and 
utilized for training neural networks. 
It is based on the principle that 
all numerical computations are ultimately compositions of a finite 
set of elementary operations for which the derivatives are known \cite{baydin}.
A chain rule combines the derivatives of the constituent operations 
and provides the derivatives of the overall composition. 

As an example for understanding the backpropagation and the AD, 
we consider a simple real-valued linear unit consisting of two nodes 
in Fig.~\ref{fig:fwbkmodel}.
Assume that a neural network containing this linear unit performs a classification task 
and its result is evaluated by loss function ${\cal L}$.
Let $W$ be a $2\!\times\!2$ real-valued matrix whose element 
$w_{jh}\!\in\!\mathbb{R}$, $j,h\!=\!1,2$, and  column vector 
${\bm x}\!=\!(x_1,x_2)^T$, $x_1,x_2\!\in\!\mathbb{R}$, 
where $(x_1,x_2)^T$ denote the transpose of $(x_1,x_2)$.
The forward process in Fig.~\ref{fig:fwbkmodel}(a) is expressed by
\begin{equation}
\begin{pmatrix} y_1\\ y_2\\\end{pmatrix}=
\begin{pmatrix}
w_{11} & w_{12}\\ w_{21} & w_{22}\\
\end{pmatrix}
\begin{pmatrix} x_1\\ x_2\\\end{pmatrix}\:.
\label{eq:fwmodel}
\end{equation}
In training of the linear unit, weight $w_{jh}$ is updated by
\begin{equation}
w_{jh}\leftarrow w_{jh}-\eta\nabla{\cal L}\;,
\end{equation}
where $\eta$ denotes the learning rate and 
$\nabla{\cal L}\!=\!\partial{\cal L}/\partial w_{jh}$ in this case.
The backward process in Fig.~\ref{fig:fwbkmodel}(b) is expressed by
\begin{align}
\begin{pmatrix} \fracslant{\partial{\cal L}}{\partial x_1}\\ 
\fracslant{\partial{\cal L}}{\partial x_2}\end{pmatrix} &=
W^T
\begin{pmatrix} \fracslant{\partial{\cal L}}{\partial y_1}\\ 
\fracslant{\partial{\cal L}}{\partial y_2}\end{pmatrix}\: ,\label{eq:bkmodel_io}\\
\frac{\partial{\cal L}}{\partial w_{jh}} &=
x_h\, \frac{\partial{\cal L}}{\partial y_j}\: .\label{eq:bkmodel_w}
\end{align}
%
Note that transformation matrices of the forward and the backward process are 
the transpose of each other.
Eqs.~(\ref{eq:bkmodel_io}) and (\ref{eq:bkmodel_w}) are derived using a chain rule
as follows.
\begin{align}
\frac{\partial{\cal L}}{\partial x_h} &= 
\sum_{j=1}^2 \frac{\partial{\cal L}}{\partial y_j}\frac{\partial y_j}{\partial x_h}
= \sum_{j=1}^2 w_{jh}\,\frac{\partial{\cal L}}{\partial y_j}\:,
\label{eq:bkmodel_io_proof}\\
\frac{\partial{\cal L}}{\partial w_{jh}} &= 
\frac{\partial{\cal L}}{\partial y_j}\frac{\partial y_j}{\partial w_{jh}}
= x_h\, \frac{\partial{\cal L}}{\partial y_j}\: .
\label{eq:bkmodel_w_proof}
\vspace*{-2.5mm}
\end{align}

\subsection{Complex-Valued Derivative}\label{subsec:complex}
In the case of a classification task,
loss function ${\cal L}$ is a real-valued function 
even in a complex-valued neural network.
We can extend the real-valued AD to complex-valued one using 
the following Wirtinger derivatives.
\begin{define}[Wirtinger derivatives \cite{hjorungnes}]
Let $f$ be a real-valued non-analytic function of $z\!\in\!\mathbb{C}$,
e.g., $f:\mathbb{C}\!\rightarrow\!\mathbb{R}$.
Let $z\!=\!\Re(z)\!+\!i\Im(z)$ and 
$z^*\!=\!\Re(z)\!-\!i\Im(z)$, 
where $i^2\!=\!-1$ and $\Re(z)$ and $\Im(z)$ are functions that return 
the real and the imaginary part of $z$,
then Wirtinger derivatives of $f$ 
with respect to $z$ and $z^*$ are defined as 
\begin{align}
\frac{\partial f}{\partial z} &= \frac{1}{2}\left( 
\frac{\partial f}{\partial\Re(z)} -i\frac{\partial f}{\partial\Im(z)}\right)
\label{eq:wir}\\
\frac{\partial f}{\partial z^*} &= \frac{1}{2}\left( 
\frac{\partial f}{\partial\Re(z)} +i\frac{\partial f}{\partial\Im(z)}\right)
\label{eq:wir_conj}\: ,
\end{align}
where $z$ and $z^*$ are regarded as independent variables \cite{brandwood}.
\label{def:wirtinger}
\end{define}
%
\noindent Note that Eq.~(\ref{eq:conj_relate}) holds from 
Eqs.~(\ref{eq:wir}) and (\ref{eq:wir_conj}).
\begin{equation}
\left( \frac{\partial f}{\partial z}\right)^* = 
\frac{\partial f}{\partial z^*} \: .
\label{eq:conj_relate}
\end{equation}

Suppose that $w_{jh},x_h\!\in\!{\mathbb C}$ in Eq.~(\ref{eq:fwmodel}).
Using the gradient descent, weight $w_{jh}$ is updated as 
\begin{equation}
w_{jh}\leftarrow w_{jh} -\eta '\,  \nabla{\cal L}\:,
\end{equation}
where $\eta '$ denotes a tentative learning rate.
The gradient of real-valued loss function ${\cal L}$ defined on complex plane
$z\!=\!\Re (z)\!+\!i\Im (z)$ is expressed by 
\begin{equation}
\nabla {\cal L} = \frac{\partial{\cal L}}{\partial\Re(w_{jh})}
	+i\frac{\partial{\cal L}}{\partial\Im(w_{jh})}
= 2\cdot\left( \frac{\partial{\cal L}}{\partial w_{jh}^*}\right)\:.
\label{eq:grad}
\end{equation}
%
Then the weight $w_{jh}$ is updated as 
\begin{equation}
w_{jh}\leftarrow w_{jh} -\eta\: \frac{\partial{\cal L}}{\partial w_{jh}^*} \:.
\end{equation}
The backward process is expressed by
\begin{align}
\begin{pmatrix} \fracslant{\partial{\cal L}}{\partial x_1^*}\\ 
\fracslant{\partial{\cal L}}{\partial x_2^*}\end{pmatrix} &=
W^{\dagger}
\begin{pmatrix} \fracslant{\partial{\cal L}}{\partial y_1^*}\\ 
\fracslant{\partial{\cal L}}{\partial y_2^*}\end{pmatrix}\: ,\label{eq:complex_bk_io}\\
\frac{\partial{\cal L}}{\partial w_{jh}^*} &=
x_h^*\, \frac{\partial{\cal L}}{\partial y_j^*}\: ,\label{eq:complex_bk_w}
\end{align}
where $W^{\dagger}$ denotes the conjugate transpose of $W$.
Thus the complex-valued derivatives are derived by using the chain rule and 
Wirtinger derivatives defined by Definition~\ref{def:wirtinger}.

\section{Proposed Learning Method}\label{sec:prop}
We propose an acceleration method for learning an fine-layered linear unit 
where a weight matrix is represented by using unitary matrices 
based on two {\em basic units} of the PSDC and the DCPS.
We first derive customized derivatives utilized in the backward process
with automatic differentiation (AD) by using the chain rule and  
Wirtinger derivatives \cite{brandwood,leung,haykin}.
Next, we show a function module implemented in C++ which 
our customized derivatives are incorporated in.

\subsection{Customized Derivatives}\label{subsec:deriv}
An MZI 
consists of a programmable PS and a DC 
with 0.5:0.5 power split ratio, two of each in our settings. 
Then there are three distinct structures of (PSDC)$^2$, (DCPS)$^2$, and 
(DCPS)(PSDC) by preventing the DC-DC sequence described in Section~\ref{subsec:mzi}.
These matrices are represented by products of two basic matrices 
of the PSDC and the DCPS.
To deal with linear units containing the MZIs with various representations,
we prepare customized functions using the two basic matrices
for the forward and the backward process in the training, 
instead of directly using the MZI-representation matrices.

\begin{figure}[t]
\begin{center}
	\subfigure[{\normalsize Forward}]{
		\psfrag{A}[c][c][1.0]{$x_1$}
		\psfrag{B}[c][c][1.0]{$x_2$}
		\psfrag{C}[c][c][1.0]{$y_1$}
		\psfrag{D}[c][c][1.0]{$y_2$}
		\psfrag{E}[c][c][1.0]{$\phi$}
		\includegraphics[width=31mm]{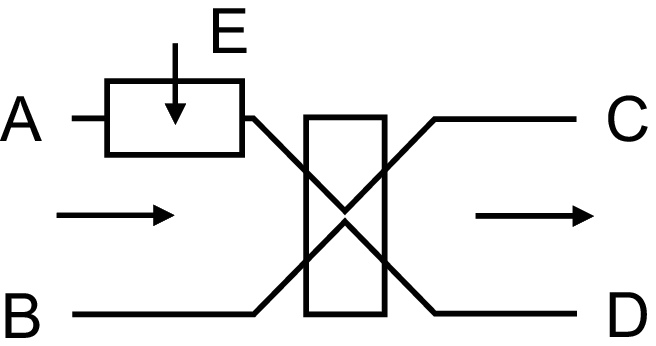}
	} \hspace*{6mm}
	\subfigure[{\normalsize Backward}]{
		\psfrag{X}[r][r][1.2]{$\frac{\partial{\cal L}}{\partial x_1^*}$}
		\psfrag{Y}[r][r][1.2]{$\frac{\partial{\cal L}}{\partial x_2^*}$}
		\psfrag{Z}[l][l][1.2]{$\frac{\partial{\cal L}}{\partial y_1^*}$}
		\psfrag{U}[l][l][1.2]{$\frac{\partial{\cal L}}{\partial y_2^*}$}
		\psfrag{V}[l][l][1.2]{$\frac{\partial{\cal L}}{\partial\phi}$}
		\includegraphics[width=31mm]{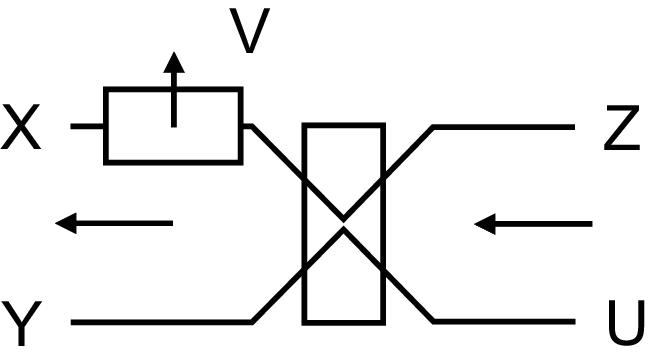}
	}
\end{center}
\vspace*{-3mm}
\caption{Diagram representing the relationship between inputs and outputs 
for (a) forward and (b) backward process in PSDC module.}
\label{fig:fwbk}
\end{figure}

Figures~\ref{fig:fwbk}(a) and (b) show diagrams representing the relationship
between the inputs and the outputs in the forward and the backward process 
in the PSDC whose structure is the same as the part of PS$_1$ and DC$_1$ 
in Fig.~\ref{fig:mzi}(b).
In the forward process, phase $\phi$ is given as an optimized parameter 
of the programmable PS besides input vector $(x_1,x_2)^T$.
In the backward process, 
$(\partial{\cal L}/\partial y_1^*,\partial{\cal L}/\partial y_2^*)^T$ is given
and $(\partial{\cal L}/\partial x_1^*,\partial{\cal L}/\partial x_2^*)^T$ 
is passed to the next layer.
For updating $\phi$ as 
$\phi\leftarrow \phi\!-\!\eta_{\phi}(\partial{\cal L}/\partial\phi)$,
derivative $(\partial{\cal L}/\partial\phi)$ is calculated, 
where $\eta_{\phi}$ denotes the learning rate of $\phi$.

\begin{proposition}
In the forward process in the PSDC, 
the linear function of the input $(x_1,x_2)^T\!\in\!{\mathbb C}^2$ is expressed by
%
\begin{equation}
\begin{pmatrix} y_1\\ y_2\\\end{pmatrix}\!=\!
\frac{1}{\sqrt{2}}
\begin{pmatrix}e^{i\phi} & i\\ ie^{i\phi}& 1\end{pmatrix}
\begin{pmatrix} x_1\\ x_2\end{pmatrix}\:.
\label{eq:fw_psdc}
\end{equation}
When Eq.~(\ref{eq:fw_psdc}) holds, in the backward process,
the input-output relationship and the derivative 
$(\partial{\cal L}/\partial\phi)$ are expressed by
\begin{equation}
\begin{pmatrix}\fracslant{\partial\cal{L}}{\partial x_1^*}\\ 
\fracslant{\partial\cal{L}}{\partial x_2^*}\end{pmatrix}\!=\!
\frac{1}{\sqrt{2}} \begin{pmatrix}
e^{-i\phi} & -ie^{-i\phi}\\-i &1\end{pmatrix}\!
\begin{pmatrix} \fracslant{\partial\cal{L}}{\partial y_1^*}\\
\fracslant{\partial\cal{L}}{\partial y_2^*}\end{pmatrix}\: ,
\label{eq:bk_psdc}
\end{equation}
\begin{equation}
\frac{\partial\cal{L}}{\partial\phi} = 
2\cdot\Im\!\left( x_1^*\frac{\partial\cal{L}}{\partial x_1^*}\right) \:.
\label{eq:bk_psdc_phi}
\end{equation}
\label{thm:psdc}
\end{proposition}
Note that the transformation matrix in the backward process is 
the conjugate transpose of that in the forward process 
and the derivative $(\partial{\cal L}/\partial\phi)$ is expressed by 
only the information passing through the interface between the phase shifter
and the other layer.
\begin{proof}
Equation~(\ref{eq:bk_psdc}) holds based on Eq.~(\ref{eq:complex_bk_io}).
By using the chain rule and Wirtinger derivatives, 
$(\partial\cal{L}/\partial\phi)$ is expressed as
\begin{align}
\frac{\partial\cal{L}}{\partial\phi}&=
\sum_{j=1}^{2}\left( 
\frac{\partial\cal{L}}{\partial y_j}\frac{\partial y_j}{\partial\phi}
\!+\!\frac{\partial\cal{L}}{\partial y_j^*}\frac{\partial y_j^*}{\partial\phi}
\right) \notag\\ &= \frac{i}{\sqrt{2}}\left\{
x_1e^{i\phi}\left(\frac{\partial\cal{L}}{\partial y_1}\!+\!
i\frac{\partial\cal{L}}{\partial y_2}\right)\!-\!
x_1^*e^{-i\phi}\left(\frac{\partial\cal{L}}{\partial y_1^*}\!-\!
i\frac{\partial\cal{L}}{\partial y_2^*}\right)\right\} \notag\\
&=i\left( x_1\frac{\partial\cal{L}}{\partial x_1}\!-\!
x_1^*\frac{\partial\cal{L}}{\partial x_1^*}\right)\notag \\
&=i\left\{ \left( x_1^*\frac{\partial\cal{L}}{\partial x_1^*} \right)^* \!-\!
x_1^*\frac{\partial\cal{L}}{\partial x_1^*}\right\}
= 2\cdot \Im\left(x_1^*\frac{\partial\cal{L}}{\partial x_1^*}\right) \: .
\label{proof:bk_psdc_phi}
\end{align}
The penultimate equality is derived by using
$(\partial{\cal L}/\partial x_1)^*\!=\! (\partial{\cal L}/\partial x_1^*)$ 
in Eq.~(\ref{eq:conj_relate}) of Wirtinger derivatives.
\end{proof}

\begin{proposition}
In the forward process in the DCPS, 
the linear function of the input $(x_1,x_2)^T\!\in\!{\mathbb C}^2$ is expressed by
%
\begin{equation}
\begin{pmatrix} y_1\\ y_2\\\end{pmatrix}\!=\!
\frac{1}{\sqrt{2}}
\begin{pmatrix}e^{i\phi} & ie^{i\phi}\\i & 1\end{pmatrix}
\begin{pmatrix} x_1\\ x_2\end{pmatrix}\: .
\label{eq:fw_dcps}
\end{equation}
When Eq.~(\ref{eq:fw_dcps}) holds, in the backward process,
the input-output relationship and the derivative 
$(\partial{\cal L}/\partial\phi)$ are expressed by
\begin{equation}
\begin{pmatrix}\fracslant{\partial\cal{L}}{\partial x_1^*}\\ 
\fracslant{\partial\cal{L}}{\partial x_2^*}\end{pmatrix}\!=\!
\frac{1}{\sqrt{2}} \begin{pmatrix}
e^{-i\phi} & -i\\-ie^{-i\phi} &1\end{pmatrix}\!
\begin{pmatrix} \fracslant{\partial\cal{L}}{\partial y_1^*}\\
\fracslant{\partial\cal{L}}{\partial y_2^*}\end{pmatrix}\: ,
\label{eq:bk_dcps}
\end{equation}
\begin{equation}
\frac{\partial\cal{L}}{\partial\phi} = 
2\cdot\Im\!\left( y_1^*\frac{\partial\cal{L}}{\partial y_1^*}\right) \: .
\label{eq:bk_dcps_phi}
\end{equation}
\label{thm:dcps}
\end{proposition}
The transformation matrices in Eqs.~(\ref{eq:fw_dcps}) and (\ref{eq:bk_dcps})
are the transpose of those in Eqs.~(\ref{eq:fw_psdc}) and (\ref{eq:bk_psdc}).
Like the PSDC, 
$(\partial{\cal L}/\partial\phi)$ is expressed by only the foregoing information.

\begin{proof}
Equation~(\ref{eq:bk_dcps}) holds based on Eq.~(\ref{eq:complex_bk_io}).
By using the chain rule and Wirtinger derivatives, 
$(\partial\cal{L}/\partial\phi)$ is expressed as
\begin{align}
\frac{\partial\cal{L}}{\partial\phi}&=
\frac{\partial\cal{L}}{\partial y_1}\frac{\partial y_1}{\partial\phi}
\!+\!\frac{\partial\cal{L}}{\partial y_1^*}\frac{\partial y_1^*}{\partial\phi}
 \notag\\ &=
\frac{i}{\sqrt{2}}\left\{\! \left( e^{i\phi}x_1\!+\!ie^{i\phi}x_2\right)
\frac{\partial\cal{L}}{\partial y_1}
\!-\!\left( e^{-i\phi}x_1^*\!-\!ie^{-i\phi}x_2\right)
\frac{\partial\cal{L}}{\partial y_1^*}\right\}\notag\\
&=i\left( y_1\frac{\partial\cal{L}}{\partial y_1}
-y_1^*\frac{\partial\cal{L}}{\partial y_1^*}\right)
= 2\cdot \Im\left(y_1^*\frac{\partial\cal{L}}{\partial y_1^*}\right)\: .
\label{proof:bk_dcpc_phi}
\end{align}
\end{proof}

By using the customized derivatives (CD) of 
Eqs.~(\ref{eq:bk_psdc_phi}) and (\ref{eq:bk_dcps_phi}),
the automatic differentiation does not need to decompose the functions to 
registered elementary functions such as an exponential function used in 
Eqs.~(\ref{eq:bk_psdc}) and (\ref{eq:bk_dcps}).
This leads to 
the acceleration of learning of the linear units 
with the PSDC and the DCPS in Section~\ref{subsec:module}.

\subsection{Function Module}\label{subsec:module}
\begin{figure}[t]
\begin{center}
	\psfrag{A}[r][r][1.0]{$x_1$} \psfrag{B}[r][r][1.0]{$x_2$}
	\psfrag{C}[r][r][1.0]{$x_3$} \psfrag{D}[r][r][1.0]{$x_4$}
	\psfrag{E}[l][l][1.0]{$y_1$} \psfrag{F}[l][l][1.0]{$y_2$}
	\psfrag{G}[l][l][1.0]{$y_3$} \psfrag{H}[l][l][1.0]{$y_4$}
	\psfrag{X}[r][r][1.0]{$\bm x$} \psfrag{Y}[l][l][1.0]{$\bm y$}
	\psfrag{I}[c][c][0.9]{$S_{A11}$} \psfrag{J}[c][c][0.9]{$S_{A12}$}
	\psfrag{K}[c][c][0.9]{$S_{B11}$} \psfrag{L}[c][c][0.9]{$S_{B12}$}
	\psfrag{M}[c][c][0.9]{$S_{A21}$} \psfrag{N}[c][c][0.9]{$S_{A22}$}
	\psfrag{O}[c][c][0.9]{$S_{B21}$} \psfrag{P}[c][c][0.9]{$S_{B22}$}
	\psfrag{Q}[c][c][0.9]{$D$}
	\includegraphics[width=83mm]{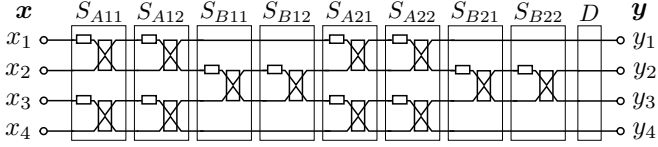}
\end{center}
\vspace*{-2mm}
\caption{
PSDC-based fine-layered structure with the rectangular structure 
that realizes any $4\!\times\! 4$ unitary matrix.
}
\label{fig:clements_psdc}
\vspace*{5mm}
\end{figure}

We design a function module that accelerates learning of a linear unit 
with {\em basic units} of the PSDC and the DCPS.
Such a linear unit with the PSDC-based fine-layered structure 
that realizes any $4\!\times\! 4$ unitary matrix
is shown in Fig.~\ref{fig:clements_psdc},
which corresponds to the MZI-based structure in Fig.~\ref{fig:clements}.
Input vector ${\bm x}$ is linearly transformed to
output vector ${\bm y}$ 
by the sequence of fine layers with unitary matrices,
$S_{A11},S_{A12},S_{B11},S_{B12},\cdots$, and $D$.
The fine layers are regularly connected with each other, that is,
the outputs in the $j$th fine layer are directly connected to the corresponding 
the inputs in the $(j\!+\!1)$th fine layer, where $1\!\leq\!j\!\leq\! 8$.
For the acceleration, 
we exploit the regularity in addition to the customized derivatives 
in Section~\ref{subsec:deriv}.

\setlength\floatsep{1pt}
\renewcommand{\baselinestretch}{1.2}
\begin{algorithm}[!t]
  \caption{Basic-unit process in linear unit} 
  \label{algo:hidden}
	\DontPrintSemicolon
	\KwIn{
		$h_{in} \in {\mathbb C}^n$, ($L$: Length of $S^{(f)}$--list)\;
	\hspace*{12mm}$S^{(f)}$--list of 
$[ S^{(f)}_{A11},S^{(f)}_{A12},S^{(f)}_{B11},S^{(f)}_{B12},\cdots]$\;
\tcp*[r]{\hspace*{-1mm}\small{$S_{\star}^{(f)}:{\mathbb C}^n\rightarrow{\mathbb C}^n$, $S_{\star}^{(f)}(h)\!=\!S_{\star}\,h$, {\rm given} $h$.}}
\tcp*[r]{\hspace*{-1mm}\small{$\star\!=\!(A11,A12,B11,B12,\cdots)$}}
	}
	\KwOut{$h_{out} \in{\mathbb C}^{n\times L}$
		\tcp*[h]{\small{$h_{out}${\rm : Collection of $h_{out(j)}$}}}
	}
	\BlankLine
	\For(in order: $j\!=\!1,\cdots,L$){$S^{(f)}_{\star}$ {\bf in} $S^{(f)}$--{\mbox{list}}}{
		$h_{out(j)}\leftarrow$ \fbox{\raisebox{0pt}[9pt][0pt]{\,$S^{(f)}_{\star}(h_{in})$}}
		\label{algo:base1}\;
		\fbox{\raisebox{0pt}[8pt][0pt]{$h_{in}\leftarrow h_{out(j)}$}}\label{algo:base2}\;
	}
	\Return{$h_{out}$}
\end{algorithm}
\renewcommand{\baselinestretch}{1.0}

Algorithm~\ref{algo:hidden} shows an overview of the basic-unit process 
in the linear unit with the fine-layered structure.
The linear unit receives 
$n$-dimensional complex-valued vector $h_{in}\!\in\!{\mathbb C}^n$
and returns the collection of $h_{out}\!\in\!{\mathbb C}^{n\times L}$,
where $L$ denotes the number of fine layers consisting of the basic units.
In each fine layer, $h_{in}$ is transformed to $h_{out(j)}$ 
by unitary matrix $S_{\star}\!\in\!{\mathbb C}^{n\times n}$, 
where $\star\!=\!A11,A12,B11,B12,\cdots$ and $j\!=\! 1,2,3,4,\cdots$ 
at line~\ref{algo:base1}
and $h_{out(j)}$ is copied to $h_{in}$ at line~\ref{algo:base2}.

In the conventional AD implemented in the Python-based machine learning frameworks,
$S_{\star}^{(f)}$ is defined only for the forward process 
at line~\ref{algo:base1} in Algorithm~\ref{algo:hidden}.
Then the AD automatically calculates values required in the backward process.
Note that lines~\ref{algo:base1} and \ref{algo:base2} are
replaced with $h_{in}\leftarrow S_{\star}^{(f)}(h_{in})$ in Python implementation.
Instead of the AD in the backward process,
we prepared functions using the customized derivatives (CD),
which were implemented as two distinct functions in Python and C++.
We call a module with the Python-implementation functions
for the forward and the backward process {\em CDpy} 
and a module with the C++-implementation functions for both the processes {\em CDcpp}.

In a function module which our proposed method is incorporated in,
we utilize the C++-implementation functions for the forward and the backward process
like {\em CDcpp}.
Furthermore, leveraging the regular connections in the fine-layered structure,
we rewire the pointer of output $h_{out(j)}$ to that of input $h_{in}$
at line~\ref{algo:base2} in Algorithm~\ref{algo:hidden}
to avoid copying the output to the input.
Since this pointer rewiring (PR) technique is exploited 
in the forward and the backward process,
the function module allows us to collectively calculate the values required in 
both the processes through all the fine layers at high speed.
The effect on speed performance is revealed in Section~\ref{subsec:res}.

Thus our function module 
has the versatility on the MZI representation and 
accelerates the forward and the backward process for learning a linear unit with
the fine-layered architecture based on the basic units of the PSDC and the DCPS.

\section{Experiments}\label{sec:exp}
We experimentally demonstrate that our proposed method worked
much faster than the conventional AD corresponding to 
the previous method in \cite{jing} 
without sacrificing accuracy.
We show our settings including 
a neural network (NN), an executed task 
for evaluating performance of a PSDC-based NN,
and parameter values for learning 
and a computer system where learning of the NN was executed, 
followed by the experimental results.

\begin{figure}[t]
\begin{center}
\psfrag{X}[r][r][0.8]{$x(t)$}
\psfrag{Q}[c][c][0.8]{${\bm h}(t)$}
\psfrag{Y}[l][l][0.8]{${\bm y}(t)$}
\psfrag{Z}[c][c][0.8]{${\bm z}(t)$}
\includegraphics[width=84mm]{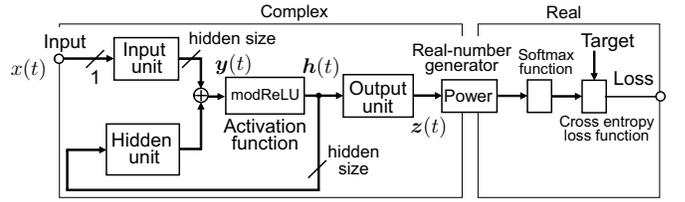}
\end{center}
\vspace*{-2mm}
\caption{Elman-type recurrent neural network for pixel-by-pixel
MNIST task where the transformation matrix in the hidden unit is 
a product of unitary matrices and diagonal unitary matrix.}
\label{fig:rnn}
\end{figure}

\subsection{Settings}\label{subsec:setting}
We employed an Elman-type recurrent neural network (RNN) model 
shown in Fig.~\ref{fig:rnn}.
The model was implemented in PyTorch 1.7.0 with C++ extension and Python 3.8.5.
With the RNN model, 
we executed a classification task using 
the MNIST handwritten digit database \cite{mnist}.
The task was pixel-by-pixel MNIST task widely used for evaluating 
performance of RNNs that contain a hidden unit with a unitary matrix 
\cite{le,arjovsky,wisdom,jing,maduranga}.
The MNIST database consists of $60,000$ training and $10,000$ test images.
Each image is $256$-level gray-scale $28\!\times\!28$ pixels 
representing a digit from $0$ to $9$.
The pixels are flattened into a sequence of $784$ pixels.
Given the pixel sequence of a single image,
the RNN sequentially receives one pixel and 
predicts a digit of the image just after the last pixel is processed.

The RNN in Fig.~\ref{fig:rnn} was composed of two blocks: 
One was a complex-number processing block and the other was 
a real-number processing block.
The complex-number processing block contained the input, the hidden,
and the output unit and the activation function. 
The hidden unit consisted of a product of unitary matrices 
with the rectangular structure and a diagonal unitary matrix
as shown in Fig.~\ref{fig:clements_psdc}.
The complex-number processing block processed an input-pixel value 
$x(t)\!\in\!{\mathbb C}$ at time $t$, 
which was generated by adding a zero imaginary part to a normalized real pixel value,
as follows.
\begin{align}
{\bm y}(t) &= \left( W_{in}\cdot x(t)+{\bm b}_{in}\right)
	+W_h\cdot {\bm h}(t\!-\! 1)\: ,\\
{\bm h}(t) &= \sigma({\bm y}(t))\: ,\\
{\bm z}(t) &= W_{out}\cdot {\bm h}(t)+{\bm b}_{out}\: ,
\label{eq:rnn}
\end{align}
where $W_{in}\!\in\!{\mathbb C}^{H\times 1}$, $W_h\!\in\!{\mathbb C}^{H\times H}$,
and $W_{out}\!\in\!{\mathbb C}^{O\times H}$ denote the weight matrices of the input,
the hidden, and the output unit.
$H$ and $O$ denote the hidden size and the output size (the number of classes).
${\bm b}_{in}\!\in\!{\mathbb C}^H$ and ${\bm b}_{out}\!\in\!{\mathbb C}^{O}$ 
denote the biases of the input and the output unit. 
$\sigma ({\bm y}(t))$ means that the following activation function called 
a modReLU function was applied to each element of ${\bm y}(t)\!\in\!{\mathbb C}^H$, 
i.e., $y_j\!\in\!{\mathbb C}$, $j\!=\!1,2,\cdots,H$, where $t$ is omitted for simplicity.
\begin{equation}
\sigma(y_j) =\left\{
\begin{array}{cc}
\frac{y_j}{|y_j|}(|y_j|+b_j) & \mbox{if}~|y_j|+b_j\geq 0\\
0 & \mbox{otherwise}
\end{array}
\right. \label{eq:modrelu} \:,
\end{equation}
where 
$b_j\!\in\!{\mathbb R}$ is an optimized bias parameter 
\cite{arjovsky},\cite{maduranga}.
We varied a hidden size and matrix representation capacity (the number of fine layers).
The hidden size ($H$) and capacity ($L$) were varied from $32$ to $1024$ 
and from $4$ to $20$.
%
The complex-valued signal ${\bm z}(t)$ passing the output unit was transformed 
to a real number in
the real-number generator whose function was power function 
$P:{\mathbb C}^O\!\rightarrow\!{\mathbb R}^O$ expressed by 
$P({\bm z}(t))\!=\!{\bm z}(t)\odot{\bm z}(t)^*$, 
which is the Hadamard product of ${\bm z}(t)$ and ${\bm z}(t)^*$.
The real-number processing block comprised the conventional units used for
a classification task.
As the loss function, a cross-entropy loss function was used.
 
We trained the RNN model with a mini-batch whose size was $100$.
Then we adopted data tensors with a feature-first structure, e.g.,
[hidden size, batch size] for the data tensor used in the hidden unit.
The feature-first tensor structure was more efficient than 
a batch-first tensor structure when a small batch size was used 
for training the RNN model in a CPU-based computer system.
For parameter optimization,
we used the RMSProp optimizer with distinct learning rates ($\eta$) for the units:
$\eta\!=\!10^{-4}$ for the input unit,
$\eta\!=\!10^{-2}$ for the output unit,
$\eta\!=\!10^{-4}$ for the hidden unit,
and $\eta\!=\!10^{-5}$ for the activation function.
The initial hidden state was fixed at zero and 
all the initial phase-shifter angles in the weight unitary matrix in the hidden unit 
were randomly sampled from $[-\pi,+\pi]$.

The pixel-by-pixel MNIST task on the foregoing model was executed 
on a computer system with Ubuntu 20.04 LTS, 
which was equipped with a single core-i7-10700K 3.8-GHz CPU
with three-level caches and a 64-GB main memory, 
by multithreading with eight threads within the memory capacity.

\subsection{Results}\label{subsec:res}
\begin{figure}[t]
\begin{center}\hspace*{1mm}
	\subfigure[{\normalsize Training accuracy}]{
		\psfrag{X}[c][c][0.87]{
			\begin{picture}(0,0)
				\put(0,0){\makebox(0,-1)[c]{Epoch}}
			\end{picture}
		}
		\psfrag{Y}[c][c][0.87]{
			\begin{picture}(0,0)
				\put(0,0){\makebox(0,25)[c]{Training accuracy}}
			\end{picture}
		}
		\psfrag{O}[c][c][0.75]{$0$}
		\psfrag{P}[c][c][0.75]{$5$}
		\psfrag{Q}[c][c][0.75]{$10$}
		\psfrag{R}[c][c][0.75]{$15$}
		\psfrag{S}[c][c][0.75]{$20$}
		\psfrag{H}[c][c][0.75]{$5$}
		\psfrag{J}[c][c][0.75]{$20$}
		\psfrag{T}[r][r][0.75]{$0.5$}
		\psfrag{U}[r][r][0.75]{$0.6$}
		\psfrag{V}[r][r][0.75]{$0.7$}
		\psfrag{W}[r][r][0.75]{$0.8$}
		\psfrag{Z}[r][r][0.75]{$0.9$}
		\psfrag{N}[r][r][0.75]{$1.0$}
		\psfrag{A}[r][r][0.70]{H32}
		\psfrag{B}[r][r][0.70]{H64}
		\psfrag{C}[r][r][0.70]{H128}
		\psfrag{D}[r][r][0.70]{H256}
		\psfrag{E}[r][r][0.70]{H512}
		\psfrag{F}[r][r][0.70]{H1024}
		\includegraphics[width=39.2mm]{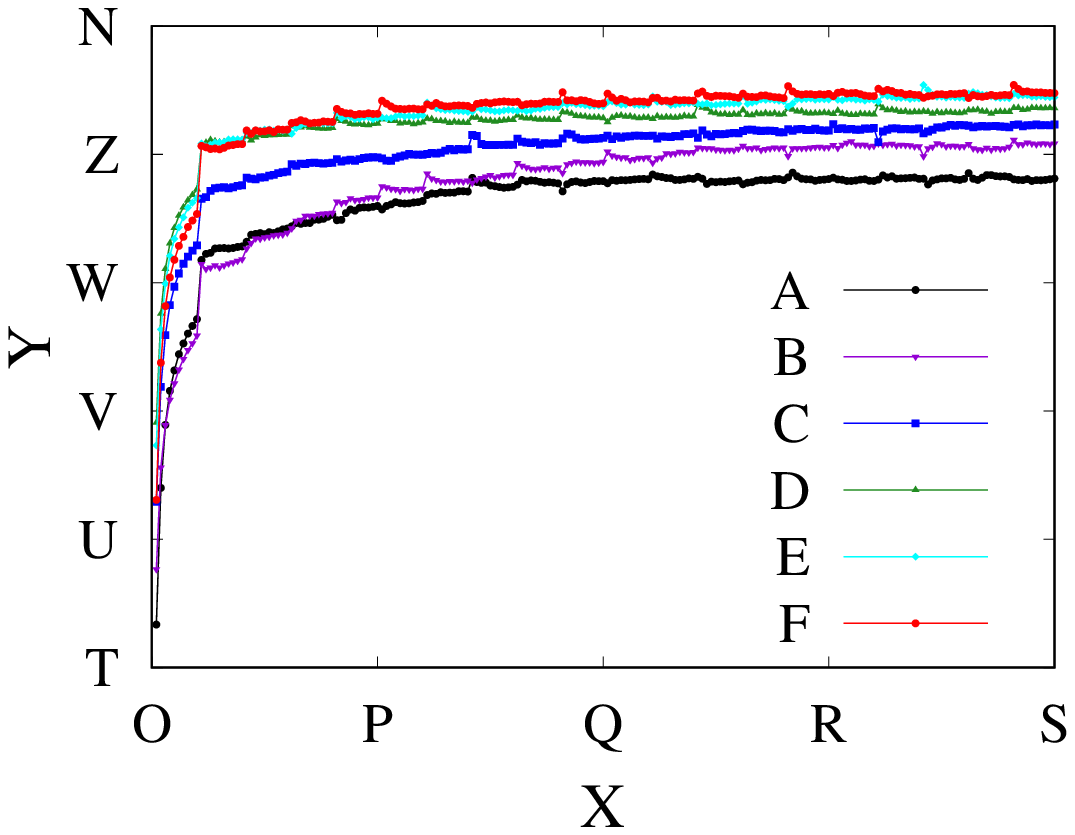}
	} \hspace*{1mm}
	\subfigure[{\normalsize Test accuracy}]{
		\psfrag{X}[c][c][0.88]{
			\begin{picture}(0,0)
				\put(0,0){\makebox(0,-5)[c]{Hidden size: H}}
			\end{picture}
		}
		\psfrag{Y}[c][c][0.87]{
			\begin{picture}(0,0)
				\put(0,0){\makebox(0,25)[c]{Test accuracy}}
			\end{picture}
		}
		\psfrag{P}[c][c][0.75]{$32$}
		\psfrag{Q}[c][c][0.75]{$64$}
		\psfrag{R}[c][c][0.75]{$128$}
		\psfrag{S}[c][c][0.75]{$256$}
		\psfrag{T}[c][c][0.75]{$512$}
		\psfrag{U}[c][c][0.75]{$1024$}
		\psfrag{W}[r][r][0.75]{$0.85$}
		\psfrag{Z}[r][r][0.75]{$0.90$}
		\psfrag{L}[r][r][0.75]{$0.95$}
		\psfrag{M}[r][r][0.75]{$1.0$}
		\psfrag{J}[c][c][0.75]{$C_1$}
		\psfrag{K}[c][c][0.75]{$C_2$}
		\psfrag{A}[r][r][0.7]{\em AD}
		\psfrag{B}[r][r][0.7]{\em Proposed}
		\includegraphics[width=39.2mm]{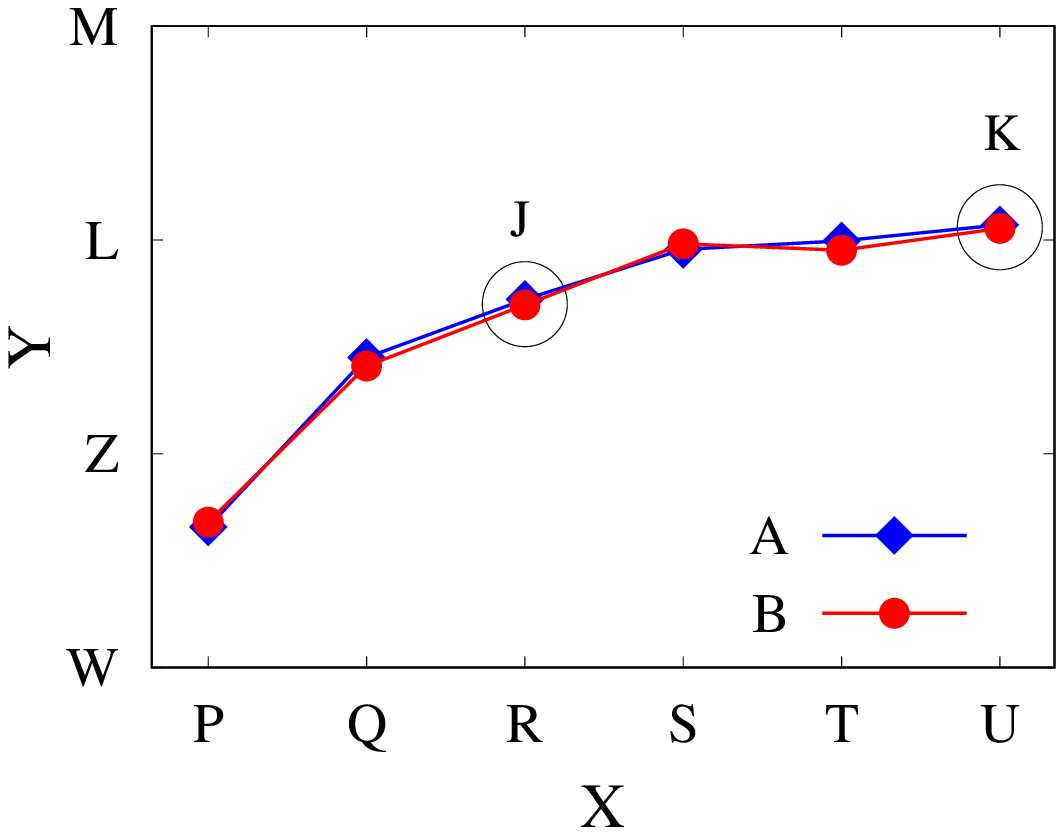}
	}
\end{center}
\vspace*{-2mm}
\caption{(a)~Training accuracy along epoch when our proposed method was 
applied to the RNN where the hidden unit had four fine layers 
and varied its size from $32$ to $1024$ (H32 to H1024).
(b)~Test accuracy of our proposed method and the conventional AD
along hidden size in log-linear scale just after $20$ epochs.
The hidden size varied from $32$ to $1024$ under the condition of 
the number of fine layers was fixed at four.
Circles $C_1$ and $C_2$ show the two test accuracies at the hidden size of 128 
related to Fig.~\ref{fig:acc_time} and at the same setting as the result in
\cite{jing}, respectively.
}
\label{fig:accuracy}
\end{figure}

We confirmed that the RNN model
was stably and successfully trained by our proposed method.
Figure~\ref{fig:accuracy}(a) shows the training accuracy along epoch
when the RNN models,
which had different hidden sizes from $32$ to $1024$ 
and the fixed number of fine layers of {\em four}, 
were trained for the pixel-by-pixel MNIST task by our method.
Note that the {\em four} fine-layer structure corresponds to 
$(S_{A11},S_{A12},S_{B11},S_{B12})$ in Fig.~\ref{fig:clements_psdc} 
equivalent to $(S_{A1},S_{B1})$ in Fig.~\ref{fig:clements} 
and the four fine layers are fewer than those necessary 
for realizing any unitary matrix.
%
%
Figure~\ref{fig:accuracy}(b) shows the test accuracy just after $20$ epochs 
along hidden size, which is displayed in log-linear scale.
The test accuracies by our method and the conventional AD were almost the same values 
and increased with the hidden size in this range.
Circle $C_2$ denotes the test accuracies at the same setting 
as that used in \cite{jing}.
Then the test-accuracy curve by the AD (PyTorch) 
can be regarded as that 
by the AD (TensorFlow 1.x) in \cite{jing}.

We demonstrate that 
our proposed method achieved much faster learning than the conventional AD that 
can be regarded as the previous method in \cite{jing}, keeping the same accuracy.
Furthermore, we present the analysis results of our method in terms of contributions 
of the constituent techniques to the speed performance.

Our two main acceleration techniques are the customized derivatives (CD)
in Section~\ref{subsec:deriv} and the collective calculation with
the output-input pointer rewiring (PR) in the function module 
in Section~\ref{subsec:module}.
To analyze the effects of the CD and the PR 
on the speed performance, 
we used four methods of the conventional AD ({\em AD}),
{\em CDpy} with the PyTorch-implementation CD for the backward process,
{\em CDcpp} with the CD implemented in C++,
and our proposed method ({\em Proposed}) 
using both the CD and the PR implemented in C++ in Section~\ref{subsec:module}.
Both the two method of {\em CDpy} and {\em CDcpp}
did not leverage the pointer rewiring (PR) technique.

\begin{figure}[t]
\begin{center}
	\psfrag{X}[c][c][0.95]{Time ($\times 10^3$ sec)}
	\psfrag{Y}[c][c][0.95]{
		\begin{picture}(0,0)
			\put(0,0){\makebox(0,20)[c]{Training accuracy}}
		\end{picture}
	}
	\psfrag{O}[c][c][0.90]{$0$}
	\psfrag{P}[c][c][0.90]{$1$}
	\psfrag{Q}[c][c][0.90]{$2$}
	\psfrag{R}[c][c][0.90]{$3$}
	\psfrag{V}[r][r][0.90]{$0.6$}
	\psfrag{W}[r][r][0.90]{$0.7$}
	\psfrag{L}[r][r][0.90]{$0.8$}
	\psfrag{M}[r][r][0.90]{$0.9$}
	\psfrag{N}[r][r][0.90]{$1.0$}
	\psfrag{A}[r][r][0.80]{\em AD}
	\psfrag{B}[r][r][0.80]{\em CDpy}
	\psfrag{C}[r][r][0.80]{\em CDcpp}
	\psfrag{D}[r][r][0.80]{\em Proposed}
	\includegraphics[width=58mm]{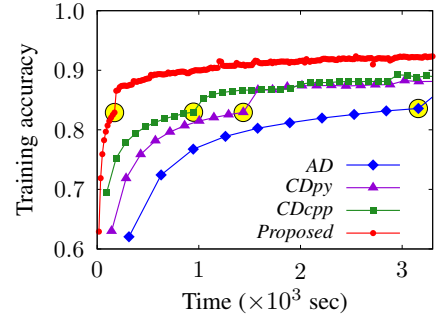}
\end{center}
\vspace*{-3mm}
\caption{Training accuracies by our proposed method ({\em Proposed}),
{\em CDcpp}, {\em CDpy}, and the conventional AD ({\em AD}) along time.
The hidden size and the number of fine layers were fixed at 
128 (H128) and four (L4).
The marks and the circle in each curve were put every $0.1$ epoch and 
1 epoch, respectively and work as indicators for measuring time.
The curves reached at almost the identical test accuracy.
In particular, {\em Proposed} and {\em AD} correspond to 
circle $C_1$ in Fig.~\ref{fig:accuracy}(b) after 20 epochs.
}
\label{fig:acc_time}
\end{figure}

Figure~\ref{fig:acc_time} shows the training accuracies of the RNN model 
by the four methods of {\em AD}, {\em CDpy}, {\em CDcpp}, and {\em Proposed},
along time until $3,300$ sec.
The RNN model had the hidden unit whose size and number of fine layers 
were fixed at 128 (H128) and four (L4).
This setting corresponds to circle $C_1$ in Fig.~\ref{fig:accuracy}(b).
The four curves were critically different in time scale 
although they reached at almost the identical test accuracy after 20 epochs.
At around $3,000$ sec, the accuracy by our method was over $0.92$ 
while that by the AD was still $0.83$.
The marks in each curve are put every $0.1$ epoch and 
the circles at one epoch work as indicators measuring the time per epoch.
Thus our method significantly saved the time required 
for training the RNN models.
This is useful to learn fine-layered neural networks using 
distinct models and many parameters under limited computing resources.

\begin{figure}[t]
\begin{center}
	\psfrag{X}[c][c][0.95]{Number of fine layers: L}
	\psfrag{Y}[c][c][0.95]{
		\begin{picture}(0,0)
			\put(0,0){\makebox(0,28)[c]{Avg. time (sec)}}
		\end{picture}
	}
	\psfrag{O}[c][c][0.90]{$0$}
	\psfrag{P}[c][c][0.90]{$4$}
	\psfrag{Q}[c][c][0.90]{$8$}
	\psfrag{R}[c][c][0.90]{$12$}
	\psfrag{S}[c][c][0.90]{$16$}
	\psfrag{T}[c][c][0.90]{$20$}
	\psfrag{U}[r][r][0.90]{$10^2$}
	\psfrag{V}[r][r][0.90]{$10^3$}
	\psfrag{W}[r][r][0.90]{$10^4$}
	\psfrag{Z}[r][r][0.90]{$5\!\times\!10^4$}
	\psfrag{M}[c][c][0.88]{$\mathbf{\displaystyle\frac{1}{19}}$}
	\psfrag{N}[c][c][0.88]{$\mathbf{\displaystyle\frac{1}{53}}$}
	\psfrag{A}[r][r][0.80]{\em AD}
	\psfrag{B}[r][r][0.80]{\em CDpy}
	\psfrag{C}[r][r][0.80]{\em CDcpp}
	\psfrag{D}[r][r][0.80]{\em Proposed}
	\includegraphics[width=58mm]{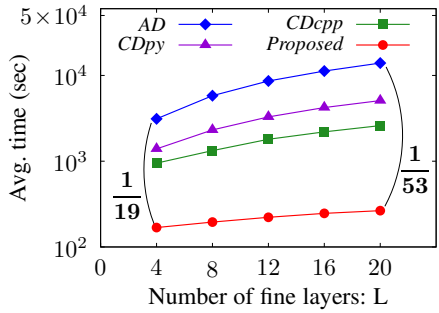}
\end{center}
\vspace*{-3mm}
\caption{
Average elapsed time per epoch (sec) along the number of fine layers 
in linear-log scale 
when the pixel-by pixel MNIST task was performed in the RNN models.
Our proposed method reduced the avg. elapsed times to ($1/19$) and ($1/53$) 
at the number of fine layers of four and 20, respectively.
}
\label{fig:time_layer}
\end{figure}

For the speed-performance analysis, 
we prepared the RNN models where 
each of their hidden units had the different number of fine layers 
from $4$ to $20$ and the fixed hidden size of $128$.
Figure~\ref{fig:time_layer} shows the average elapsed time per epoch (sec) 
of the four methods,
{\em AD}, {\em CDpy}, {\em CDcpp}, and {\em Proposed}, 
along the number of fine layers in linear-log scale
when the RNN models were trained for the pixel-by-pixel MNIST task in $20$ epochs.
When the number of fine layers was four, 
the test accuracies of {\em AD} and {\em Proposed} 
correspond to those at circle $C_1$ in Fig.~\ref{fig:accuracy}(b) 
and the learning curves of the four methods are illustrated in Fig.~\ref{fig:acc_time}.
In terms of speed performance, 
our proposed method worked $19$ and $53$ times faster than {\em AD}
at the number of fine layers of $4$ and $20$.
{\em CDpy} and {\em CDcpp} performed about twice and $4$ times 
the acceleration from {\em AD}.
The remaining acceleration effect came from the pointer-rewiring (PR) technique 
implemented in C++,
which collectively calculates the values used in the forward and the backward process.
The function module is a simple yet very effective technique.
Thus our proposed method accelerated 
the learning of the RNN model 
whose hidden unit was constructed with the fine-layered liner unit. 

\section{Conclusion}\label{sec:conc}
We proposed an acceleration method for learning parameters 
of Mach-Zehnder interferometers (MZIs) in a fine-layered linear unit
in an optical neural network (ONN).
Our proposed method employed a function module in C++
to collectively calculate values of customized complex-valued 
derivatives for a product of unitary matrices 
representing an MZI in the linear unit.
Consequently, our method reduces the time required for learning
the parameters.

We confirmed that 
it worked almost 20 times faster than the conventional 
automatic differentiation (AD)
when a pixel-by-pixel MNIST task was performed in 
a complex-valued recurrent neural network 
with an MZI-based hidden unit.
Since our method is compatible with the current AD,
we can easily use the method in 
machine learning platforms.

The two directions remain as future work.
One is to implement our proposed method in codes available in
the state-of-the-art computer systems such as those equipped with 
multiple GPUs.
The other is to compare ours with the other methods using 
linear units with unitary matrices as their weights like 
those in \cite{maduranga}, \cite{mhammedi}  beyond the usage for ONNs
and explore how to apply our individual techniques to them.



\end{document}